\documentclass[10pt,twocolumn,letterpaper]{article}

\usepackage[pagenumbers]{cvpr} 

\definecolor{cvprblue}{rgb}{0.21,0.49,0.74}
\usepackage[pagebackref,breaklinks,colorlinks,allcolors=cvprblue]{hyperref}
\usepackage{array}
\usepackage{multirow}
\usepackage{graphicx}
\usepackage{amsmath}
\usepackage{amssymb}
\usepackage{pifont}

\def\paperID{*****} 
\def\confName{CVPR}
\def\confYear{2025}

\title{FollowGen: A Scaled Noise Conditional Diffusion Model for Car-Following Trajectory Prediction}

\author{
    Junwei You, Rui Gan, Weizhe Tang, Zilin Huang, Jiaxi Liu, Zhuoyu Jiang, Haotian Shi\textsuperscript{\dag}, \\ Keshu Wu, Keke Long, Sicheng Fu, Sikai Chen\textsuperscript{\dag}, Bin Ran
}

\begin{document}
\maketitle

\begingroup
\renewcommand\thefootnote{}
\footnotetext{\textsuperscript{\dag}Corresponding Authors.}
\endgroup

\begin{abstract}
Vehicle trajectory prediction is crucial for advancing autonomous driving and advanced driver assistance systems (ADAS). Although deep learning-based approaches—especially those utilizing transformer-based and generative models—have markedly improved prediction accuracy by capturing complex, non-linear patterns in vehicle dynamics and traffic interactions, they frequently overlook detailed car-following behaviors and the inter-vehicle interactions critical for real-world driving applications, particularly in fully autonomous or mixed traffic scenarios. To address the issue, this study introduces a scaled noise conditional diffusion model for car-following trajectory prediction, which integrates detailed inter-vehicular interactions and car-following dynamics into a generative framework, improving both the accuracy and plausibility of predicted trajectories. The model utilizes a novel pipeline to capture historical vehicle dynamics by scaling noise with encoded historical features within the diffusion process. Particularly, it employs a cross-attention-based transformer architecture to model intricate inter-vehicle dependencies, effectively guiding the denoising process and enhancing prediction accuracy. Experimental results on diverse real-world driving scenarios demonstrate the state-of-the-art performance and robustness of the proposed method.

\end{abstract}

\section{Introduction}

\begin{figure}[htbp]
\centering
\includegraphics[width=.5\textwidth]{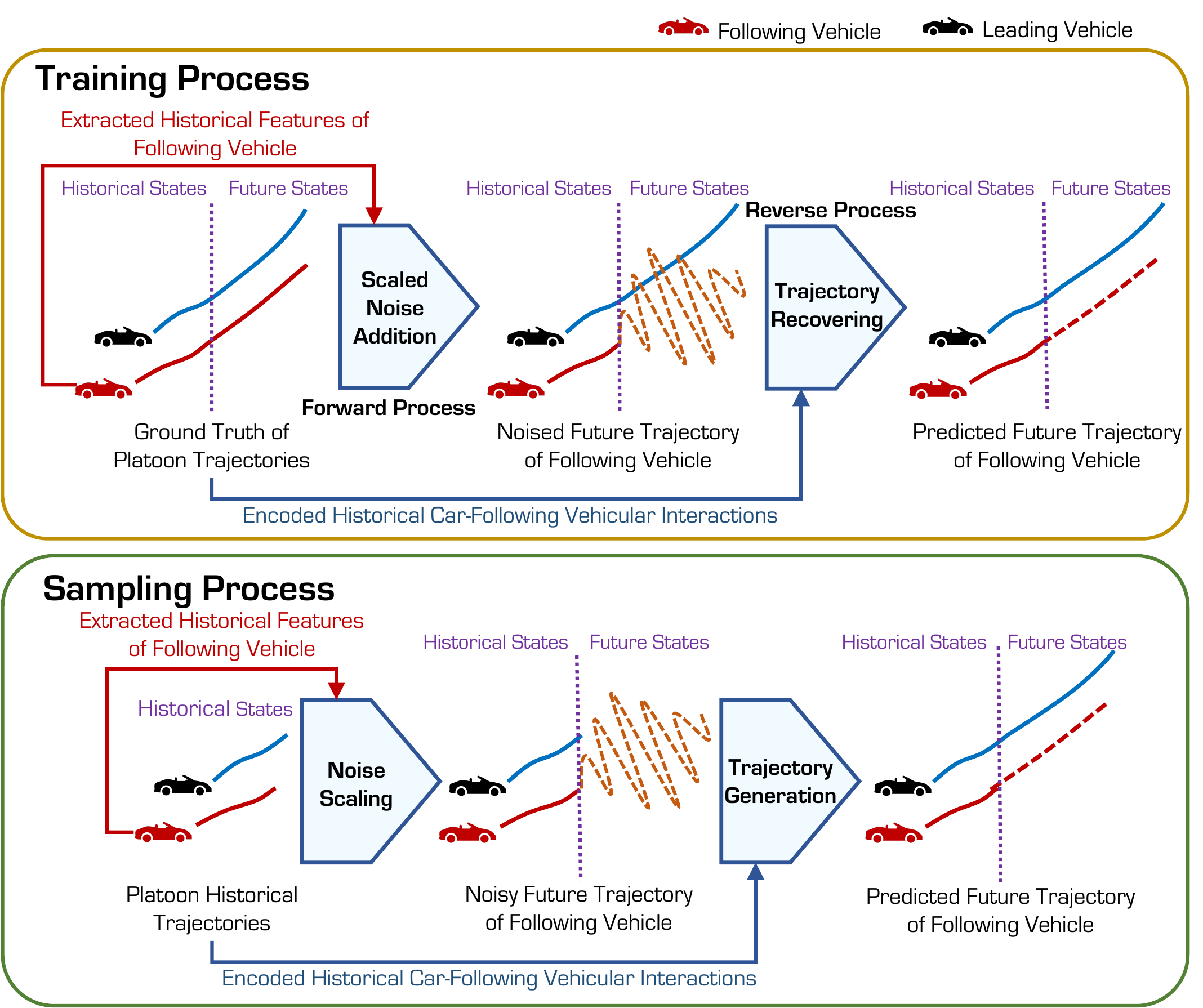}
\caption{The proposed FollowGen model is a generative framework that predicts the future trajectory of a following vehicle by modeling its historical interactions with a leading vehicle. During the training process, historical features of the following vehicle's states are extracted to scale the noise that is added progressively to the ground truth of its future trajectory. The model learns a reverse process to recover the original trajectory, where the fine-grained platoon inter-vehicle interactions are encoded to guide the denoising process. In the sampling process, the model generates future trajectories by sampling from the noise scaled by historical features effectively. where the denoising process is directed by the learned car-following interactions.}
\label{fig:yourlabel}
\end{figure}

Vehicle trajectory prediction is critical in advancing autonomous driving and advanced driver assistance systems (ADAS). Predicting future vehicle positions is essential for collision avoidance, route planning, and adaptive cruise control. In recent years, deep learning-based approaches have propelled the field forward by providing adaptive solutions that can learn complex, nonlinear patterns and temporal dependencies from large datasets \cite{altche2017lstm, dai2019modeling, benterki2019long, long2024physics, nikhil2018convolutional, xie2020motion, mo2020interaction, wang2023vehicle, chen2021s2tnet, chen2022vehicle, geng2023physics, mo2021graph, chen2021spatial, lu2022vehicle, wu2023graph, sadid2024dynamic, SHENG2024100142}. Among them, denoising diffusion model-based generative models have attracted great attention for their ability to capture diverse motion patterns under uncertainty. Recent advancements in these models provide a flexible framework for generating multimodal trajectory distributions and refining noise into realistic motion sequences \cite{wei2024diff, jiang2023motiondiffuser, cheng2023gatraj, chen2023equidiff}. By incorporating spatial constraints, social interactions, and geometric properties, diffusion-based models have the potential to produce vehicle trajectories that closely align with real-world dynamics, which makes them highly effective for autonomous driving applications where accurate predictions are critical.

To further improve prediction accuracy, recent studies integrate interaction information between agents within deep learning models, focusing on general interactions with surrounding vehicles to capture the complex dynamics between a target vehicle and neighboring agents in dynamic traffic scenarios.  For instance, the study \cite{guo2022end} proposes a grid-based method, which utilizes occupancy grid maps centered on the target vehicle and surrounding agents to approximate trajectory distributions. Extending grid-based techniques, the hierarchical CapsNet framework \cite{qin2023spatiotemporal} uses geographic grid maps to encode both spatial and temporal dependencies among vehicles, preserving the spatial relationships essential for predictions in Vehicle-to-Everything (V2X) networks. Additionally, graph-based models combine GCNs with temporal encoders or attention mechanisms to capture multi-agent interactions, jointly predicting vehicle trajectories with interpretability \cite{sheng2022graph, schmidt2022crat, zhang2022ai, xu2023agnp}.

In addition to general interactions, car-following behavior has traditionally been a central focus in traffic modeling due to its direct impact on safe following distances and the stability of traffic flow. Recently, integrating car-following dynamics into deep learning-based trajectory prediction models has gained momentum, enabling more precise modeling of inter-vehicle dependencies. For instance, TransFollower \cite{zhu2022transfollower} introduces a transformer-based architecture that leverages historical driving data and future speed profiles of the leading vehicle to capture the dependencies in car-following behavior. Another study \cite{shi2022integrated} proposes an integrated prediction framework that combines car-following and lane-changing behaviors in a unified framework with BiLSTM and TCN layers, along with an attention-based switch mechanism, to capture transitions between these behaviors, thus enabling precise trajectory predictions in dynamic traffic environments.

Inspired by prior research on car-following behavior, it is clear that the impact of detailed car-following dynamics on trajectory prediction merits further investigation. Unlike broader multi-agent interactions, car-following behavior focuses on fine-grained adjustments, such as acceleration, deceleration, and precise following distance relative to the leading vehicle, which is essential, particularly in dense traffic scenarios. These micro-interactions are also important in fully autonomous vehicles (AVs) or mixed environments, where precise responses to the leading vehicle are essential for maintaining safe and efficient traffic flow \cite{huang2024human}. In particular, in mixed-traffic environments where AVs and human-driven vehicles (HVs) coexist, AVs must account for the variability and unpredictability of the leading HV behaviors. However, trajectory prediction based specifically on car-following behaviors remains largely unexplored in both mixed and AV-only contexts, leaving a gap in current deep-learning models that do not fully capture this critical information. At the same time, diffusion model-based generative frameworks offer a promising but unexplored approach for capturing the dynamics of car-following behavior in trajectory prediction. 

Hence, this study aims to develop a diffusion model-based generative framework for predicting vehicle trajectory integrating the detailed car-following interactions between two adjacent vehicles in varied traffic environments, as illustrated in Fig. 1. The main contributions of this paper are as follows:
\begin{itemize}
\item We introduce FollowGen, a novel generative framework for forecasting vehicle trajectory in car-following scenarios incorporating detailed inter-vehicle interactions.

\item We develop a temporal feature encoding pipeline consisting of GRU, vehicle location-based attention, and Fourier embedding, effectively extracting the temporal features from the historical vehicle trajectory.

\item We propose a noise scaling strategy that conditions the isotropic Gaussian noise on encoded historical movement features of the vehicle. Scaled noise substitutes isotropic noise in the diffusion process. 


\item We model the car-following inter-vehicle dynamics via a cross-attention-based transformer architecture. The extracted interaction embedding is induced in the denoising network to guide the trajectory generation process.


\item We validate the robustness and generality of  FollowGen on multiple real-world scenarios, including HV following HV, HV following AV, and AV following HV, through comparative and ablative studies. 
\end{itemize}

\section{Related Work}

\textbf{Generative Model-Based Trajectory Prediction.} Generative models effectively enhance vehicle trajectory prediction by capturing the inherent uncertainty and variability in driving behavior that deterministic models often overlook. These models can generate a distribution of possible future trajectories, producing a more comprehensive and realistic prediction framework. The major types of generative models include GANs, flow-based methods, VAEs, and diffusion models. In general, GAN-based models utilize a generator to produce plausible trajectories and a discriminator to evaluate their realism, which refines predictions through adversarial training \cite{choi2021dsa, chen2022cae, zhao2019multi, wang2020multi}. Unlike GANs, flow-based approaches transform a simple distribution into a complex one by learning invertible mappings and thus enable the generation of diverse trajectories \cite{ma2020diverse, chen2024mixed}. VAEs and their variants typically encode trajectories into a latent space and decode them back, allowing the generation of diverse trajectories by sampling from the latent space \cite{de2022vehicles, lee2022muse, feng2019vehicle, neumeier2021variational, kim2021driving}. 

\textbf{Diffusion Models in Trajectory Prediction.} Diffusion models have recently gained prominence in trajectory prediction due to their robust ability to handle uncertainty and generate diverse, realistic trajectories. Starting from a simple, usually Gaussian distribution, diffusion models gradually transform this distribution into the complex distribution of real-world trajectories by learning the underlying data structure. Moreover, diffusion models have shown superiority in handling complex traffic scenarios where interactions between multiple agents and environment must be carefully considered. The flexibility of the diffusion framework to incorporate spatial and temporal dependencies through advanced architectures, such as transformers and GNNs, further enhance their predictive performance. In summary, diffusion models' capability to learn and generalize from large datasets with robustness to uncertainty makes them a prevailing method to predict trajectories. 

Plenty of studies in literature have verified this. A diffusion-based model for environment-aware trajectory prediction is introduced in \cite{westny2024diffusion}, where its robustness and ability to handle complex traffic scenarios by leveraging conditional diffusion processes to model trajectory uncertainty is highlighted. A multi-modal vehicle trajectory prediction framework presented in \cite{li2023multi} uses a conditional diffusion model to address trajectory sparsity and irregularity in world coordinate systems. Combined with CNNs, a hierarchical vector transformer diffusion model developed in \cite{tang2024hierarchical} captures trajectory uncertainty and further improves prediction performance. Another trajectory prediction framework called motion indeterminacy diffusion (MID) introduced in \cite{gu2022stochastic}, is designed to handle the indeterminacy of human behavior and provide accurate stochastic trajectory predictions. A recent study \cite{tang2024utilizing} has also explored using a diffusion model for pedestrian trajectory prediction in semi-open autonomous driving environments, focusing on reducing computational overhead and improving the accuracy of multi-agent joint trajectory predictions. The Conditional Equivariant Diffusion Model (EquiDiff) \cite{chen2023equidiff} combines the diffusion model with SO(2) equivariant transformer to utilize the geometric properties of location coordinates. It also applies RNNs and Graph Attention Networks (GAT) to extract social interactions from historical trajectories. Although the majority of diffusion-based trajectory prediction models have incorporated social interactions into their structures for improved performance, the integration of detailed car-following behaviors and complex vehicular interactions remains unexplored. Therefore, the proposed FollowGen is dedicated to integrating the advantages of diffusion models with finer details of car-following dynamics to capture detailed vehicle adjustments and enhance trajectory prediction performance.

\section{Methodology}
\subsection{Problem Formulation}

To formalize the trajectory prediction problem in the context of car following, we first group vehicles as platoons. For simplicity, if a platoon contains two vehicles where one vehicle is driving followed by another, the vehicle in the front is defined as the leading vehicle while the other one is the following vehicle. FollowGen aims to capture the intricate dynamics and the probabilistic nature of inter-vehicular dependencies within a platoon. Let $\mathbf{x}_i^{\rm his} \in \mathbb{R}^{T_{\text{his}} \times D}$ and $\mathbf{v}_i^{\rm his} \in \mathbb{R}^{T_{\text{his}} \times 1}$ denote the historical positions and speeds of the $i$-th vehicle in a two-vehicle platoon, where $T_{\text{his}}$ is the number of historical time steps, $D$ is the spatial dimensionality (e.g., $D=2$ for 2D positions), and $i \in \{\rm lea, \rm fol\}$ denotes the leading and following vehicles respectively. Let $\Delta\mathbf{x}^{\rm his} \in \mathbb{R}^{T_{\text{his}} \times D}$ and $\Delta\mathbf{v}^{\rm his} \in \mathbb{R}^{T_{\text{his}} \times 1}$ represent the historical spacing and speed difference between the two vehicles. The goal is to predict the future trajectory $\hat{\mathbf{x}}_{\rm fol}^{\rm fut} \in \mathbb{R}^{T_{\text{fut}} \times D}$ of the following vehicle given the historical information stated above, as shown in the equation below:

\begin{equation}
    \hat{\mathbf{x}}_{\rm fol}^{\rm fut} = f(\mathbf{x}_{\rm fol}^{\rm his}, \mathbf{v}_{\rm fol}^{\rm his}, \mathbf{x}_{\rm lea}^{\rm his}, \mathbf{v}_{\rm lea}^{\rm his}, \Delta\mathbf{x}^{\rm his}, \Delta\mathbf{v}^{\rm his})
\end{equation}
where $f(\cdot)$ represents the proposed FollowGen model. The overall framework of FollowGen is shown in Fig. 2. The proposed model has four main modules: Historical Feature Encoding (Section 3.2.1), which uses a structured pipeline to encode the following vehicle's historical features, Noise Scaling and Addition (Section 3.2.2), which scales Gaussian noise with historical features; Car-Following Interaction Modeling (Section 3.3.1), which captures detailed car-following dependencies; and Condition Guided Denoising (Section 3.3.2) which removes noise using a denoising network guided by car-following interactions for efficient trajectory prediction.
\begin{figure*}[htbp]
\centering
\includegraphics[width=\textwidth]{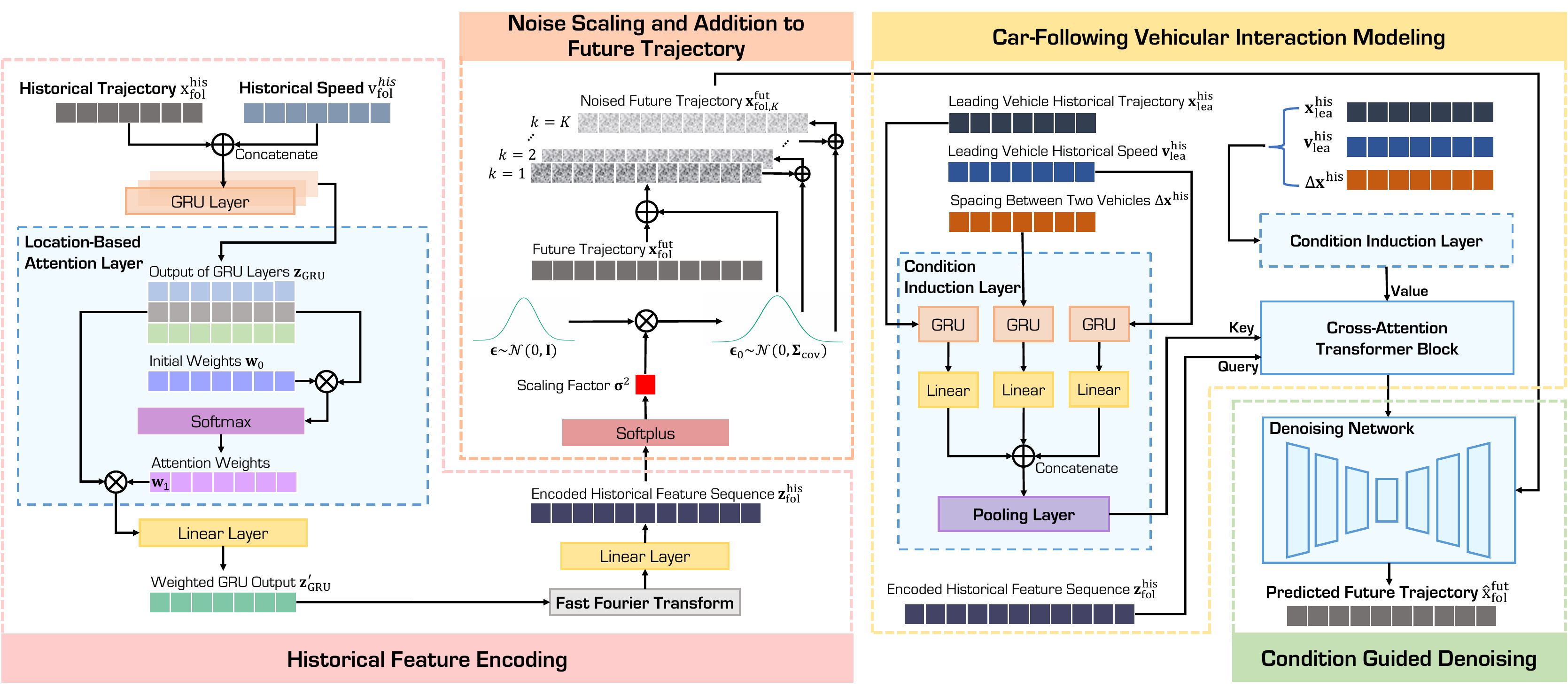}
\caption{\textbf{Overview  of FollowGen.} The proposed model consists of four main modules: Historical Feature Encoding, which uses a pipeline of GRU layers, a location-based attention layer, and FFT to encode the historical features of the following vehicle; Noise Scaling and Addition, where the Gaussian noise is scaled by the encoded historical features and progressively added to the future trajectory; Car-Following Vehicular Interaction Modeling, which captures the intricate car-following dynamics through Condition Induction Layers and a cross-attention transformer block; and Condition Guided Denoising, which leverages a denoising network to remove noise gradually and produce a refined trajectory prediction. This denoising is guided by the modeled car-following interactions, which enables the model to generate accurate future trajectories, taking advantage of detailed vehicle interaction patterns.}
\label{fig:yourlabel}
\end{figure*}

\subsection{Forward Process}
\subsubsection{Historical Feature Encoding}

We design a pipeline to encode effective features from historical trajectories. Initially, the historical trajectory $\mathbf{x}_{\rm fol}^{\rm his} \in \mathbb{R}^{T_{\text{his}} \times D}$ of the following vehicle is concatenated with its historical speed $\mathbf{v}_{\rm fol}^{\rm his} \in \mathbb{R}^{T_{\text{his}} \times 1}$, forming an input $\mathbf{z}_{\text{input}} \in \mathbb{R}^{T_{\text{his}} \times (D+1)}$, which is then fed into stacked GRU layers \cite{dey2017gate} for data fusion and temporal feature extraction. The output of the GRU layers $\mathbf{z}_{\rm GRU} \in \mathbb{R}^{T_{\text{his}} \times H}$ is then passed through a location-based attention layer.

Given the initial attention weights $\mathbf{w}_0 \in \mathbb{R}^{T_{\text{his}} \times 1}$, the operation of the location-based attention is formulated as follows:
\begin{equation} 
    \mathbf{w}_1 = \text{softmax}\left( \mathbf{W} \cdot (\mathbf{z}_{\rm GRU} \odot \mathbf{w}_0) + \mathbf{b} \right)
\end{equation}
\begin{equation} 
    \mathbf{z}_{\rm loc} =  \mathbf{w}_1 \odot \mathbf{z}_{\rm GRU}
\end{equation}
where $\mathbf{W} \in \mathbb{R}^{H \times 1}$ is the weight matrix for linear projection, $\mathbf{b} \in \mathbb{R}^{1}$ is a bias vector, $\mathbf{w}_1 \in \mathbb{R}^{T_{\text{his}} \times 1}$ represents the updated attention weights, $\mathbf{z}_{\rm loc} \in \mathbb{R}^{T_{\text{his}} \times H}$ is the output of the location-based attention layer, and $\odot$ refers to Hadamard product. Subsequently, after another linear projection, the GRU output $\mathbf{z}_{\rm GRU}$ is transformed into a weighted sequence denoted as $\mathbf{z}_{\rm GRU}^{'} \in \mathbb{R}^{T_{\text{his}} \times H'}$.

While GRU is adept at capturing long-term temporal dependencies of a sequence, Fast Fourier transform (FFT) \cite{duhamel1990fast, zhang2023learnable} is applied subsequently to reveal the periodic patterns. Taking the weighted GRU output $\mathbf{z}_{\rm GRU}^{'}$ as input, FFT is formulated as the equation below:
\begin{equation} 
    \mathbf{z}_{\rm FFT}[i] = \sum_{n=0}^{N-1} \mathbf{z}_{\rm GRU}^{'}[n] \cdot e^{-j \cdot 2 \pi \cdot \frac{i \cdot n}{N}}
\end{equation}
where $N$ is the length of the input sequence, $\mathbf{z}_{\rm GRU}^{'}[n]$ is the value of the input time-domain sequence at the $n$-th sample, $n \in [0,N-1]$, $e^{-j \cdot 2 \pi \cdot \frac{i \cdot n}{N}}$ is the complex exponential function that represents the basis functions of FFT, $j$ is the imaginary unit, and $\mathbf{z}_{\rm FFT}[i]$ is the value of the transformed frequency-domain sequence at the $i$-th frequency bin, $i \in [0,N-1]$. A linear layer is connected at the end, yielding the encoded historical feature embeddings $\mathbf{z}_{\rm fol}^{\rm his}$. 

\subsubsection{Noising Scaling and Addition}

Generally, the noise scaling strategy reshapes the isotropic Gaussian noise used in the diffusion process by extracting historical trajectory features. By conditioning the noise on historical features, we ensure that the forward diffusion process incorporates necessary conditions or restrictions reflective of the system's true dynamics. This results in a more informed and directed process of transitioning from data to noise, ensuring that the generated future trajectories are not only a product of random noise but are informed by the system's past. Statistically, this means that instead of sampling from the standard normal distribution $\mathcal{N}(0, \mathbf{I})$ as what traditional diffusion models would do, the noise is now sampled from the distribution $\mathcal{N}(0, \mathbf{\Sigma}_{\rm cov})$, where the covariance matrix $\mathbf{\Sigma}_{\rm cov} \in \mathbb{R}^{H'' \times H''}$ is a diagonal matrix, and is represented from the encoded historical embedding $\mathbf{z}_{\rm fol}^{\rm his} \in \mathbb{R}^{T_{\text{his}} \times H''}$. Specifically, to find $\mathbf{\Sigma}_{\rm cov}$, we first take the mean along the time dimension of $\mathbf{z}_{\rm fol}^{\rm his}$ which yields a vector denoted as $\mathbf{\mu} \in \mathbb{R}^{H''}$, and then apply the Softplus activation function \cite{zheng2015improving} upon $\mathbf{\mu}$. The resulting vector, denoted as $\mathbf{\sigma}^2 \in \mathbb{R}^{H''}$, is used as the scaling factor to reshape the standard normal distribution to maintain the variance of $\mathbf{\Sigma}_{\rm cov}$. This process is formulated as follows:
\begin{equation} 
    \mathbf{\sigma}^2 = \log(1 + e^{\mathbf{\mu}})
\end{equation}
\begin{equation}
    \mathbf{\Sigma}_{\rm cov} = \text{diag}(\mathbf{\sigma}^2)
\end{equation}

In practice, given that $\mathbf{\epsilon}_0 \sim \mathcal{N}(0, \mathbf{I})$ is an independent standard normal variable randomly sampled from a standard normal distribution, the scaled noise denoted as $\mathbf{\epsilon}$ can also be expressed directly as follows:
\begin{equation}
   \mathbf{\epsilon} = \mathbf{\Sigma}_{\rm cov}^{\frac{1}{2}}\mathbf{\epsilon}_0 = \text{diag}(\mathbf{\sigma}) \cdot \mathbf{\epsilon}_0, \quad \mathbf{\epsilon} \sim \mathcal{N}(0, \mathbf{\Sigma_{\rm cov}})
\end{equation}

On this basis, the forward incremental noise addition process will take the future trajectory $\mathbf{x}^{\rm fut}_{\rm fol}$ as input and gradually add the scaled noise to the input for $K$ time steps, which is formulated as follows:
\begin{equation}
    \mathbf{x}_{{\rm fol},k}^{\rm fut} = \sqrt{\alpha_k} \mathbf{x}_{{\rm fol},k-1}^{\rm fut} + \sqrt{\beta_k} \mathbf{\epsilon}, \quad \mathbf{\epsilon} \sim \mathcal{N}(0, \mathbf{\Sigma}_{\rm cov})
\end{equation}
\begin{equation}
    q(\mathbf{x}_{{\rm fol},k}^{\rm fut} | \mathbf{x}_{{\rm fol},k-1}^{\rm fut}) = \mathcal{N}(\mathbf{x}_{{\rm fol},k}^{\rm fut}; \sqrt{\alpha_k} \mathbf{x}_{{\rm fol},k-1}^{\rm fut}, \beta_k \mathbf{\Sigma}_{\rm cov})
\end{equation}
where $\beta_k$ is the time step-specific factor to control the intensity of the noise added at each step, $\mathbf{\epsilon}$ represents the noise vector sampled from a Gaussian distribution with covariance matrix $\mathbf{\Sigma}_{\rm cov}$ as stated above, $\mathbf{x}_{{\rm fol},k}^{\rm fut}$ is the data distribution at time step $k$ after undergoing $k$ times of noise addition, $\mathbf{x}_{{\rm fol},k-1}^{\rm fut}$ is the data vector at the previous time step $k-1$, and $\alpha_k = 1 - \beta_k$. 

Define $\bar{\alpha}_k = \prod_{i=1}^{k} \alpha_i$, and the diffusion process at any step $k$ from the original data $\mathbf{x}_{\rm fol}^{\rm fut}$ can be expressed in a closed form:
\begin{equation}
\mathbf{x}_{{\rm fol},k}^{\rm fut} = \sqrt{\bar{\alpha}_k} \mathbf{x}_{\rm fol}^{\rm fut} + \sqrt{(1 - \bar{\alpha}_k)} \mathbf{\epsilon}, \quad \mathbf{\epsilon} \sim \mathcal{N}(0, \mathbf{\Sigma}_{\rm cov})
\end{equation}
\begin{equation}
q(\mathbf{x}_{{\rm fol},k}^{\rm fut} | \mathbf{x}_{\rm sfol}^{\rm fut}) = \mathcal{N}(\mathbf{x}_{{\rm fol},k}^{\rm fut}; \sqrt{\bar{\alpha}_k} \mathbf{x}_{\rm fol}^{\rm fut}, (1 - \bar{\alpha}_k) \mathbf{\Sigma}_{\rm cov})
\end{equation}
Ultimately, when  $K \xrightarrow{} \infty$, $\mathbf{x}_{{\rm fol},K}^{\rm fut}$ will approximate following the prior noise distribution used in the diffusion process, $\mathbf{x}_{{\rm fol},K}^{\rm fut} \sim \mathcal{N}(0, \mathbf{\Sigma}_{\rm cov})$. The distribution of the entire sequence from $\mathbf{x}_{{\rm fol}}^{\rm fut}$ to $\mathbf{x}_{{\rm fol},K}^{\rm fut}$ conditioned on the original data $\mathbf{x}_{{\rm fol}}^{\rm fut}$ is shown as follows:
\begin{equation}
    q(\mathbf{x}_{{\rm fol},1:K}^{\rm fut} |  \mathbf{x}_{{\rm fol}}^{\rm fut}) = \prod_{k=1}^{K} q( \mathbf{x}_{{\rm fol},k}^{\rm fut} |  \mathbf{x}_{{\rm fol},k-1}^{\rm fut})
\end{equation}

\subsection{Reverse Process}
\subsubsection{Car-Following Vehicular Interaction Modeling}

We propose a cross-attention-based transformer architecture \cite{gheini2021cross} to model the intricate dependencies and dynamic interactions between the specified car-following variables. As mentioned in Section 3.1, these variables, including historical positions and speed profiles of leading vehicles, as well as the spacing and speed difference, are processed through dedicated GRU layers to capture temporal patterns. The results are then linearized and concatenated to pass through a pooling layer, which forms the key and value vectors, denoted as $\mathbf{K}$ and $\mathbf{V}$, respectively, where $\mathbf{K} \in \mathbb{R}^{d_k}$, $\mathbf{V} \in \mathbb{R}^{d_v}$. We can also define the following vehicle's encoded trajectory $\mathbf{z}_{\rm fol}^{\rm his}$ as a query vector $\mathbf{Q}$, $\mathbf{Q} \in \mathbb{R}^{d_q}$. We can also define the following vehicle's encoded trajectory $\mathbf{z}_{\rm fol}^{\rm his}$ as a query vector $\mathbf{Q}$, $\mathbf{Q} \in \mathbb{R}^{d_q}$. $\mathbf{Q}$ is used within the cross-attention transformer block to selectively weigh the leading vehicles' features, synthesizing a contextualized output that incorporates the interactive behavior of the vehicles in a platoon. The formulation of $\mathbf{Q}$, $\mathbf{K}$ and $\mathbf{V}$ is shown as follows:

\begin{equation}
\mathbf{Q} = \mathbf{z}_{\rm stu}^{\rm his}
\end{equation}
\begin{equation}
    \begin{split}
        \mathbf{K, V} = \text{Pooling} (\text{Concat} (
        & \text{Linear}(\text{GRU}(\mathbf{x}_{\rm lea}^{\rm his})), \\
        & \text{Linear}(\text{GRU}(\mathbf{v}_{\rm lea}^{\rm his})), \\
        & \text{Linear}(\text{GRU}(\Delta \mathbf{x}^{\rm his})))
    \end{split}
\end{equation}

Expressing the multi-head cross-attention operation as:
\begin{equation}
    \mathbf{z}_{\rm MCA} = \text{Concat}(\mathbf{head}_1, ..., \mathbf{head}_i, ..., \mathbf{head}_h) \mathbf{W}^{\rm out}
\end{equation}
where $\mathbf{W}^{\rm out}$ is the output weight matrix that linearly transforms the concatenated vector from all the heads into the desired output dimension, each attention head, $\mathbf{head}_i$, is computed as the equation below:
\begin{equation}
    \mathbf{head}_i = \text{softmax}\left(\frac{\mathbf{Q}\mathbf{W}_i^{\rm que} (\mathbf{K}\mathbf{W}_i^{\rm key})^{\rm T}}{\sqrt{d_k}}\right) \cdot (\mathbf{W}_i^{\rm val})
\end{equation}
where $\mathbf{W}_i^{\rm que}$, $\mathbf{W}_i^{\rm key}$, and $\mathbf{W}_i^{\rm val}$ are the parameter matrices specific to each head for the queries, keys, and values.

Finally, the output of the cross-attention transformer block enriched with vehicle relational information is embedded into a denoising network to direct and enhance the prediction accuracy. 

\subsubsection{Condition Guided Denoising}

\begin{table*}[h!]
\centering
\caption{Comparison of performance metrics for various scenarios and methods. $T$ denotes the prediction horizon.} 
\resizebox{\textwidth}{!}{%
\begin{tabular}{c|c|c|cccc|cccc|cccc}
\hline
\multirow{3}{*}{\begin{tabular}[c]{@{}c@{}}Scenario\\ \\ \end{tabular}} & \multirow{3}{*}{\begin{tabular}[c]{@{}c@{}}Method\\ \\ \end{tabular}} & \multirow{3}{*}{\begin{tabular}[c]{@{}c@{}}Reference\\ \\ \end{tabular}} & \multicolumn{4}{c|}{$T = 3s$} & \multicolumn{4}{c|}{$T = 4s$} & \multicolumn{4}{c}{$T = 5s$} \\ \cline{4-15} 
 &  &  & \textbf{RMSE $\downarrow$} & \textbf{ADE $\downarrow$} & \textbf{FDE $\downarrow$} & \textbf{MR $\downarrow$} & \textbf{RMSE $\downarrow$} & \textbf{ADE $\downarrow$} & \textbf{FDE $\downarrow$} & \textbf{MR $\downarrow$} & \textbf{RMSE $\downarrow$} & \textbf{ADE $\downarrow$} & \textbf{FDE $\downarrow$} & \textbf{MR $\downarrow$} \\ \hline
\hline
\multirow{4}{*}{H-H} & BAT \cite{liao2024bat} & AAAI 2024 & 4.0164 & 2.5038 & 5.1758 & 0.6305 & 5.5283 & 3.4799 & 7.9652 & 0.8524 & 7.8580 & 4.7988 & 12.6530 & 0.9055 \\  
 &TUTR \cite{shi2023trajectory} & ICCV 2023 & 2.7052 & \textbf{1.1174} & \textbf{1.6702} & 0.3383 & 3.3790 & \textbf{1.4093} & \textbf{2.3401} & 0.4207 & 4.3609 & 2.0374 & 3.9805 &  0.6359\\  
 & CRAT-Pred \cite{9811637} & ICRA 2022 & \textbf{2.0206} & 1.3937 & 1.8320 & \textbf{0.3368} & \textbf{2.2785} & 1.5880 & 2.5610 & 0.5200 & \textbf{2.7573} & \textbf{1.8666} & 3.6184 & 0.6630 \\  
 & \textbf{FollowGen (Ours)} & - & 2.8001 & 1.6162 & 2.1796 & 0.3820 & 3.3270 &1.7993 & 2.4480 & \textbf{0.3930} & 3.8935 & 1.9853 & \textbf{3.3454} & \textbf{0.4935} \\ \hline\hline

\multirow{4}{*}{A-H} & BAT \cite{liao2024bat} & AAAI 2024 & 2.5197 & 1.7636 & 3.3817 & 0.5412 & 3.5547 & 2.4021 & 5.3486 & 0.7403 & 5.0211 & 3.2552 & 8.0671 & 0.8434 \\  
 &TUTR \cite{shi2023trajectory} & ICCV 2023 & 2.7680 & 1.7002 & 2.1110 & 0.3977 & 2.8772 & 2.1057 & 1.9980 & 0.5017 & 4.0043 & 2.4040 & 7.4894 & 0.6469 \\  
 & CRAT-Pred \cite{9811637}& ICRA 2022 & 3.3064 & 2.0385 & 2.3917 & 0.4610 & 3.3852 & 2.1548 & 2.8396 & 0.5296 & 3.5344 & 2.2977 & 3.4711 & 0.6167 \\  
 & \textbf{FollowGen (Ours)} & - & \textbf{2.0033} & \textbf{1.3058} & \textbf{1.3243} & \textbf{0.1891} & \textbf{2.1220} & \textbf{1.3585} & \textbf{1.5601} & \textbf{0.2547} & \textbf{2.4108} & \textbf{1.5058} & \textbf{3.3469} & \textbf{0.5750} \\ \hline\hline

\multirow{4}{*}{H-A} & BAT \cite{liao2024bat} & AAAI 2024 & \textbf{1.7281} & 1.3641 & 2.4452 & 0.4922 & 2.3706 & 1.7790 & 3.7898 & 0.6646 & 3.3827 & 2.3755 & 6.1516 & 0.7583 \\  
 & TUTR \cite{shi2023trajectory}& ICCV 2023 & 2.3891 & 1.4175 & 1.8991 & \textbf{0.3191} & 2.4693 & 1.5028 & \textbf{1.7316} & \textbf{0.3355} & \textbf{2.2395} & 2.0727 & 4.2551 &  0.5334\\  
 & CRAT-Pred \cite{9811637} & ICRA 2022 & 3.7319 & 1.8575 & 1.8340 & 0.3261 & 3.6453 & 1.8801 & 2.1347 & 0.3842 & 3.7715 & 1.9899 & 3.1889 & \textbf{0.4751} \\  
 & \textbf{FollowGen (Ours)} & - & 1.9550 & \textbf{1.3218} & \textbf{1.7257} & 0.3289 & \textbf{2.1989} & \textbf{1.4516} & 1.9596 & 0.3758 & 2.4810 & \textbf{1.5970} & \textbf{2.5775} & 0.4863 \\ \hline
\end{tabular}%
}
\end{table*}

Taking the scaled noise $\mathbf{\epsilon} \sim \mathcal{N}(0, \mathbf{\Sigma}_{\rm cov})$ as input, the denoising network reconstructs the corresponding clean and accurate future trajectory of the following vehicle. In doing so, it reverses the diffusion process by sequentially predicting and removing the noise distribution introduced at each time step, thereby progressively restoring the corresponding trajectory to its original uncorrupted distribution. Given the estimated data distribution $\mathbf{x}_{{\rm fol},k}^{\rm fut}$ at any time step $k$ and the contextual information $\mathbf{c}$ from the cross-attention transformer block, estimation of the data distribution at time $k-1$ is shown as the following equation:
\begin{equation}
    p_\theta(\mathbf{x}_{{\rm fol},k-1}^{\rm fut}| \mathbf{x}_{{\rm fol},k}^{\rm fut}, \mathbf{c}) = \mathcal{N}\left(\mathbf{x}_{{\rm fol},k-1}^{\rm fut} ; \mathbf{\mu}_\theta(\mathbf{x}_{{\rm fol},k}^{\rm fut}, k, \mathbf{c}), \mathbf{\Sigma}_\theta(k)\right)
\end{equation}
where $\mathbf{\mu}_\theta\big( \mathbf{x}_{{\rm fol},k}^{\rm fut},\, k,\, \mathbf{c} \big) \in \mathbb{R}^{T_{\text{fut}} \times D}$ is the predicted mean for recovering $\mathbf{x}_{{\rm fol},k-1}^{\rm fut}$, informed by the context encoding $\mathbf{c}$, and $\mathbf{\Sigma}_\theta(k) \in \mathbb{R}^{T_{\text{fut}} \times D \times D}$ is the learned covariance matrix at time step $k$. The joint probability over the sequence conditioned on  $\mathbf{c}$, is given by:
\begin{equation}
    p_\theta(\mathbf{x}_{{\rm fol},0:K}^{\rm fut} | \mathbf{c}) = p(\mathbf{x}_{{\rm fol},K}^{\rm fut}) \prod_{k=1}^{K} p_\theta(\mathbf{x}_{{\rm fol},k-1}^{\rm fut} | \mathbf{x}_{{\rm fol},k}^{\rm fut}, \mathbf{c})
\end{equation}
\begin{equation}
    p(\mathbf{x}_{{\rm fol},K}^{\rm fut}) = \mathcal{N}(\mathbf{x}_{{\rm fol},K}^{\rm fut}; 0, \mathbf{\Sigma}_{\rm cov})
\end{equation}

\subsection{Training Objective}

The training objective of FollowGen contains three parts. The first is to maximize the variational lower bound (ELBO), which can be simplified to measure the accuracy of noise prediction, as shown below:
\begin{equation}
    \mathcal{L}_{\text{simp}}(\theta) = \mathbb{E}_{\mathbf{x}_{{\rm fol}}^{\rm fut},\ \mathbf{\epsilon} \in \mathbb{R}^{T_{\text{fut}} \times D},\ k}\left[ \left\| \mathbf{\epsilon} - \hat{\mathbf{\epsilon}}_\theta(\mathbf{x}_{{\rm fol},k}^{\rm fut},\ k,\ \mathbf{c}) \right\|^2 \right]
\end{equation}



In the second part, to discourage unrealistic predictions where the following vehicle overtakes the leading vehicle, we introduce a spacing penalty. Denote the projected longitudinal spacing between the leading vehicle's future trajectory and the following vehicle's predicted trajectory as:
\begin{equation}
    \Delta x^{\text{fut}} = \left( \mathbf{x}_{\text{lead}}^{\text{fut}} - \hat{\mathbf{x}}_{\text{fol}}^{\text{fut}} \right)^\top \mathbf{e}_{\text{d}},
\end{equation}
where $\mathbf{e}_{\text{d}}$ is the unit vector in the direction of travel. The spacing penalty $\mathcal{L}_{\text{spacing}}$ is defined as:
\begin{equation}
    \mathcal{L}_{\text{spacing}} = \mathbb{E} \left[
    \begin{cases}
        0, & \text{if } \Delta x^{\text{fut}} \geq 0, \\
        \tfrac{1}{2} \left( -\Delta x^{\text{fut}} \right)^2, & \text{if } -\delta < \Delta x^{\text{fut}} < 0, \\
        \delta \left( -\Delta x^{\text{fut}} - \tfrac{1}{2} \delta \right), & \text{if } \Delta x^{\text{fut}} \leq -\delta,
    \end{cases}
    \right]
\end{equation}
where $\delta$ is a threshold parameter controlling the transition between the quadratic and linear penalty regions. This formulation penalizes predictions where the following vehicle is ahead of the leading vehicle and thus ensues realistic spacing patterns.

The third part applies a collision penalty to prevent the following vehicle from getting too close to the leading vehicle by enforcing a minimum rational safe distance:

\begin{equation}
    \mathcal{L}_{\text{collision}} = \mathbb{E}\left[ \exp\left(-\frac{\Delta \mathbf{x}^{\text{fut}}}{dist}\right) \right]
\end{equation}
where $dist$ is a predefined threshold for safe spacing.

The total loss function, \(\mathcal{L}_{\text{total}}\), combines these three components with $\lambda_1$ and $\lambda_2$ as weighting factors for the spacing and collision penalties:

\begin{equation}
    \mathcal{L}_{\text{total}} = \mathcal{L}_{\text{simp}} + \lambda_1 \mathcal{L}_{\text{spacing}} + \lambda_2 \mathcal{L}_{\text{collision}}
\end{equation}

In the denoising network, the noise at each time step is predicted through an adapted U-Net \cite{ronneberger2015u} that incorporates the encoded car-following interactions. 




\section{Experiments}

\subsection{Experimental Setup}
\textbf{Dataset.} We evaluate the proposed FollowGen on the Large Car-Following Dataset Based on the Lyft Level-5 Dataset \cite{li2023large}. This dataset includes over 29,000 HV-following-AV (H-A), 9,000 AV-following-HV (A-H), and 42,000 HV-following-HV (H-H) instances, with a total driving distance of more than 150,000 kilometers, providing diverse car-following scenarios essential for our study. 

\textbf{Metrics.} We evaluate our model using Root Mean Squared Error (RMSE), Average Displacement Error (ADE), Final Displacement Error (FDE), and Missing Rate (MR) to test the trajectory prediction performance across various time horizons. RMSE captures the overall prediction accuracy by averaging errors across all timestamps, while FDE measures the Euclidean distance between predicted and actual endpoints, which offers an assessment of final position accuracy. ADE averages the displacement errors across each timestamp, reflecting cumulative accuracy over the trajectory. Finally, MR calculates the percentage of predictions exceeding a specified FDE threshold of 2 meters, which provides insight into prediction reliability across different horizons. In this study, each car-following scenario is evaluated separately. 

\begin{figure*}[htbp]
\centering
\includegraphics[width=0.88\textwidth]{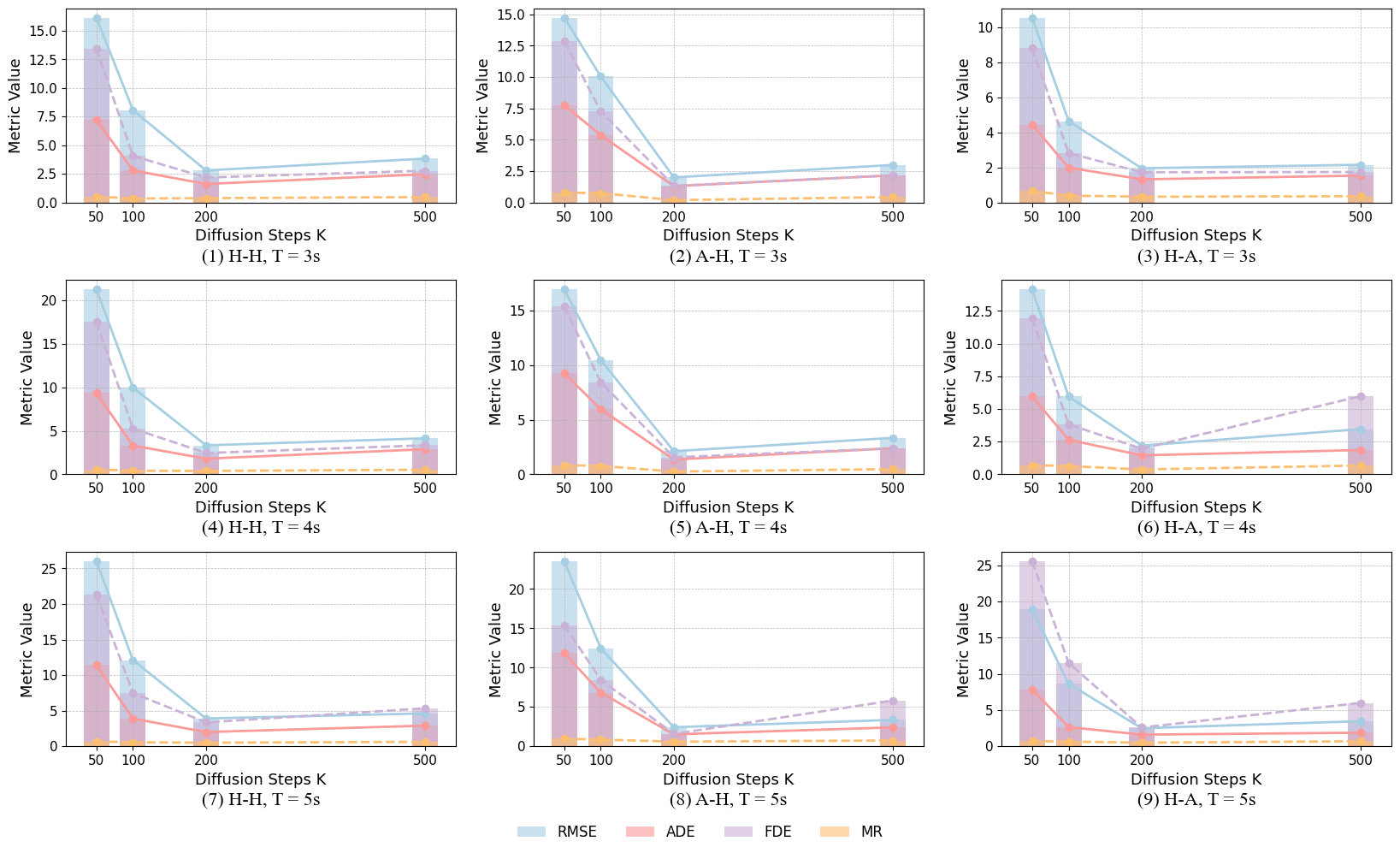}
\caption{Performance on metrics at various diffusion steps \( K \) for different car-following scenarios, over three prediction horizons.}
\label{fig:yourlabel}
\end{figure*}

\begin{table*}[h!]
\centering
\caption{Performance comparison using different beta schedules.}
\resizebox{\textwidth}{!}{%
\begin{tabular}{c|c|cccc|cccc|cccc}
\hline
\multirow{3}{*}{\begin{tabular}[c]{@{}c@{}}Scenario\\ \\ \end{tabular}} & \multirow{3}{*}{\begin{tabular}[c]{@{}c@{}}Beta Schedule \\ \\ \end{tabular}} & \multicolumn{4}{c|}{$T = 3s$} & \multicolumn{4}{c|}{$T = 3s$} & \multicolumn{4}{c}{$T = 3s$} \\ \cline{3-14} 
 &  & \textbf{RMSE $\downarrow$} & \textbf{ADE $\downarrow$} & \textbf{FDE $\downarrow$} & \textbf{MR $\downarrow$} & \textbf{RMSE $\downarrow$} & \textbf{ADE $\downarrow$} & \textbf{FDE $\downarrow$} & \textbf{MR $\downarrow$} & \textbf{RMSE $\downarrow$} & \textbf{ADE $\downarrow$} & \textbf{FDE $\downarrow$} & \textbf{MR $\downarrow$} \\ \hline
\multirow{4}{*}{H-H} 
 & Sigmoid & 4.2851 & 1.5325 & 2.4972 & \textbf{0.2956} & 5.4935 & 1.8874 & 3.3699 & 0.4096 & 6.8005 & 2.2915 & 4.5075 & 0.5047 \\ 
 & Quadratic  & 8.4458 & 2.6966 & 4.7719 & 0.3266 & 11.0898 & 3.4134 & 6.2121 & 0.4104 & 13.6338 & 4.1467 & 8.3684 & 0.5555 \\ 
 & \textbf{Linear} & \textbf{2.8001} & \textbf{1.6162} & \textbf{2.1796} & 0.3820 & \textbf{3.3270} & \textbf{1.7993} & \textbf{2.4480} & \textbf{0.3930} & \textbf{3.8935} & \textbf{1.9853} & \textbf{3.3454} & \textbf{0.4935} \\ \hline\hline
\multirow{4}{*}{A-H} 
 & Sigmoid & 3.0649 & 1.7227 & 2.0861 & 0.2922 & 3.5841 & 1.9092 & 2.8009 & 0.4422 & 5.8921 & 2.3884 & 11.0662 & 0.7297 \\ 
 & Quadratic  & 5.3219 & 2.5167 & 3.6401 & 0.3391 & 6.6435 & 2.9143 & 4.5345 & 0.4766 & 7.2045 & 3.3058 & 5.9278 & 0.7156 \\ 
 & \textbf{Linear} & \textbf{2.0033} & \textbf{1.3058} & \textbf{1.3243} & \textbf{0.1891} & \textbf{2.1220} & \textbf{1.3585} & \textbf{1.5601} & \textbf{0.2547} & \textbf{2.4108} & \textbf{1.5058} & \textbf{3.3469} & \textbf{0.5750} \\ \hline\hline
\multirow{4}{*}{H-A} 
 & Sigmoid & 3.2790 & 1.4667 & 2.3978 & 0.3102 & 4.3634 & 1.8063 & 3.2070 & 0.4074 & 11.5774 & 2.7165 & 28.5375 & 0.7264  \\ 
 & Quadratic & 7.3878 & 2.3232 & 4.0280 & \textbf{0.2599} & 9.5611 & 2.9146 & 5.2464 & \textbf{0.3639} & 12.3935 & 3.7146 & 15.2788 & 0.7370 \\ 
 & \textbf{Linear} & \textbf{1.9550} & \textbf{1.3218} & \textbf{1.7257} & 0.3289 & \textbf{2.1989} & \textbf{1.4516} & \textbf{1.9596} & 0.3758 & \textbf{2.4810} & \textbf{1.5970} & \textbf{2.5775} & \textbf{0.4863} \\ \hline
\end{tabular}%
}
\end{table*}

\textbf{Implementation Details.} The proposed FollowGen is trained with Intel Core I7 CPUs and a single NVIDIA RTX 4090 GPU. For each scenario of H-A, A-H, and H-H, we employ a diffusion process with steps $K=200$ using a linear beta schedule. The augmented loss function $\mathcal{L}_{\text{total}}$ is applied with weights $\lambda_1 = \lambda_2 = 0.001$ based on hyperparameters. $\delta$ is set to be 2, and $dist$ is defined as 2 meters. AdamW is applied as the optimizer with a learning rate of 0.001 and an epsilon value of 0.01. With the batch size of 64, the training process converges substantially within 20 epochs for both HA and HH cases, while it takes 50 epochs to converge for the AH scenario. Gradient clipping with a maximum norm of 1.0 was applied for stability.

\subsection{Comparison with State-of-the-art}

\begin{table*}[h!]
\centering
\caption{Ablation study for each car-following scenario over a five-second prediction horizon.}
\resizebox{\textwidth}{!}{%
\begin{tabular}{c|cccc|cccc}
\hline
\multirow{2}{*}{Scenario} & \multicolumn{4}{c|}{Components} & \multicolumn{4}{c}{Metrics ($T = 5s$)} \\ \cline{2-9} 
 & \textbf{Location-Based Attention} & \textbf{FFT} & \textbf{Noise Scaling} & \textbf{Cross-Attention Transformer} & \textbf{RMSE $\downarrow$} & \textbf{ADE $\downarrow$} & \textbf{FDE $\downarrow$} & \textbf{MR $\downarrow$} \\ \hline
\multirow{3}{*}{H-H} &  &  & \ding{51} & \ding{51} & 9.2043 & 2.9689 & 5.0611 & 0.5690 \\ 
 & \ding{51} & \ding{51} &  & \ding{51} & 5.5091 & 2.6537 & 4.8104 & 0.6542  \\ 
 & \ding{51} & \ding{51} & \ding{51} & & 9.2043 & 2.9689 & 5.0611 & 0.5690 \\
 & \ding{51} & \ding{51} & \ding{51} & \ding{51} &  \textbf{3.8935} & \textbf{1.9853} & \textbf{3.3454} & \textbf{0.4935}  \\ \hline\hline
 
\multirow{3}{*}{A-H} &  &  & \ding{51} & \ding{51} & 6.9448 & 2.2482 &  15.2104& 0.8766 \\ 
 & \ding{51} & \ding{51} &  & \ding{51} & 13.3004 & 4.1724 & 6.9859 &  0.7719\\ 
 & \ding{51} & \ding{51} & \ding{51} & & 7.0572 & 3.7236 & 10.7793 & 0.7828  \\
 & \ding{51} & \ding{51} & \ding{51} & \ding{51} &  \textbf{2.4108} & \textbf{1.5058} & \textbf{3.3469} & \textbf{0.5750}  \\ \hline\hline
 
\multirow{3}{*}{H-A} &  &  & \ding{51} & \ding{51} & 6.3557 & 2.4066 & 6.2522 & 0.7687 \\ 
 & \ding{51} & \ding{51} &  & \ding{51} & 5.6177 & 2.4793 & 12.8238 & 0.7329 \\ 
 & \ding{51} & \ding{51} & \ding{51} &  & 20.5695 & 8.8842 & 11.7999 & 0.6578\\
 & \ding{51} & \ding{51} & \ding{51} & \ding{51} &  \textbf{2.4810} & \textbf{1.5970} & \textbf{2.5775} & \textbf{0.4863}  \\ \hline
\end{tabular}%
}
\end{table*}

We compare the proposed FollowGen model with state-of-the-art methods, including BAT~\cite{liao2024bat}, TUTR~\cite{shi2023trajectory}, and CRA-T-Pred~\cite{9811637}, across three car-following scenarios: H-H, A-H, and H-A, evaluated over prediction horizons of \( T = 3s, 4s, \text{and } 5s \), as presented in Table 1. FollowGen demonstrates superior prediction capabilities, which is particularly evident in metrics that emphasize final position accuracy and reliability, such as FDE and MR. For instance, in the H-H scenario, at \( T = 5s \), FollowGen achieves an FDE of 3.35, which is approximately 44\% lower than the 5.18 of BAT and 11\% lower than the 3.62 of CRA-T-Pred. Similarly, the MR is 42\% lower than the 0.91 of BAT and 25\% lower than the 0.66 of CRA-T-Pred, showcasing FollowGen's robustness in predicting human-human interactions. In the A-H scenario, FollowGen's FDE at \( T = 5s \) is 3.35, compared to 8.07 for BAT, translating to a 59\% improvement. MR is also significantly reduced by 32\% compared to TUTR. The H-A scenario sees similar benefits: FollowGen achieves an FDE of 2.58 at \( T = 5s \), which is 24\% percent lower than the 3.19 of CRA-T-Pred. The MR is also 20\% lower than the value achieved by BAT, which reflects FollowGen's strength in mixed human-AV interactions. These findings suggest that FollowGen’s diffusion-based generative approach adeptly captures the complexities and stochastic nature of interactions in diverse scenarios, especially in mixed-traffic flows.

\subsection{Diffusion Process Evaluation}

We examine the effect of the diffusion process parameters on FollowGen’s performance to further understand its advantages. Specifically, we evaluate the impact of varying diffusion steps \( K \) and beta schedules.

In this study, we tested four values of diffusion steps: \( K = 50, 100, 200, \text{ and } 500 \). The results are compared in Fig. 2. A small number of steps, such as \( K = 50 \), result in insufficient refinement of the generated trajectories, and leading to less accurate predictions. Conversely,  excessive steps, like \( K = 500 \), increase computational cost without significant improvement in performance and may cause over-smoothing. We found that \( K = 200 \) achieves a balanced level of refinement, efficiently capturing essential driving behaviors. The choice of beta schedule also influences performance, with the linear schedule consistently outperforming the sigmoid and quadratic options, as shown in Table 2. This suggests that the linear schedule enables smoother noise variance control during diffusion, which enhances the prediction stability, especially in mixed human-AV interaction scenarios where uncertainty is high.

\subsection{Ablation Study}

An ablation study is conducted to evaluate the contribution of each component in the proposed FollowGen model. We construct three variants of the FollowGen model by systematically removing or altering key components and assessing their impact on prediction performance. 
The logic of constructing these variants is described as follows: 

\textbf{w/o Noise Scaling.} This variant avoids using the scaling factor encoded from historical information; instead, it still uses the isotropic Gaussian noise during diffusion.

\textbf{w/o Location-Based Attention and FFT.} This variant eliminates the GRU layers, the location-based attention, and the FFT layer used for historical information encoding. A linear layer is used as a substitute.

\textbf{w/o Cross-Attention Transformer.} This variant similarly replaces the cross-attention transformer block with a linear layer for car-following dependency modeling. 

The results of the ablation study are shown in Table 3. For all scenarios, the full model configuration consistently achieves the best performance across all metrics, which verifies the importance of each component in FollowGen. Notably, the absence of noise scaling or cross-attention transformer, key elements in capturing historical and car-following dependencies, leads to a marked degradation in predictive accuracy. In the A-H scenario, for instance, removing the cross-attention transformer increases the FDE from 3.35 to 15.21, highlighting its critical role in modeling dependencies. Similarly, in the H-A scenario, excluding noise scaling causes the FDE to rise from 2.58 to 11.80, demonstrating the importance of noise scaling in adapting to historical information for robust predictions. The results suggest that each component contributes uniquely to enhancing the model's capability to capture the complex dynamics of vehicle interactions, and their synergy is crucial for accurate and reliable predictions.

\section{Conclusion}
This study develops FollowGen, a novel approach for vehicle car-following trajectory prediction through a scaled noise conditional diffusion model. Experimental results demonstrate that the model better predicts vehicle trajectories across diverse real-world scenarios by encoding historical features and integrating detailed inter-vehicular car-following dynamics within the generative framework. Future work could extend FollowGen to more complex interactions like lane-changing, integrate multimodal sensor data for adaptability, and optimize diffusion efficiency for potential real-time applications.

{
    \fontsize{8pt}{11pt}\selectfont 
    \bibliographystyle{ieeenat_fullname}
    \bibliography{main}
}

\clearpage
\setcounter{page}{1}
\maketitlesupplementary


\section{Key Module Architectures}

The section presents the architectures of the cross-attention transformer block and the denoising network, the critical modules of FollowGen that are not detailed earlier. 

\subsection{Cross-Attention Transformer Block}
The cross-attention transformer block \cite{gheini2021cross}, shown in Fig. 4, models inter-vehicle dependencies using \(\mathbf{Q}\), \(\mathbf{K}\), and \(\mathbf{V}\) derived from historical and car-following features. The block employs scaled dot-product attention, where \(\mathbf{Q}\) interacts with \(\mathbf{K}\) through a normalized softmax function, producing attention weights. These weights are used to combine \(\mathbf{V}\) into a context-rich representation. Multi-head attention ensures diverse interaction patterns are captured, and the outputs are concatenated and transformed through a linear layer. Residual connections and layer normalization stabilize training, resulting in the refined output \(\mathbf{z}_{\text{CAT}}\) that encapsulates fine-grained vehicle interactions.

\begin{figure}[htbp]
\centering
\includegraphics[width=.5\textwidth]{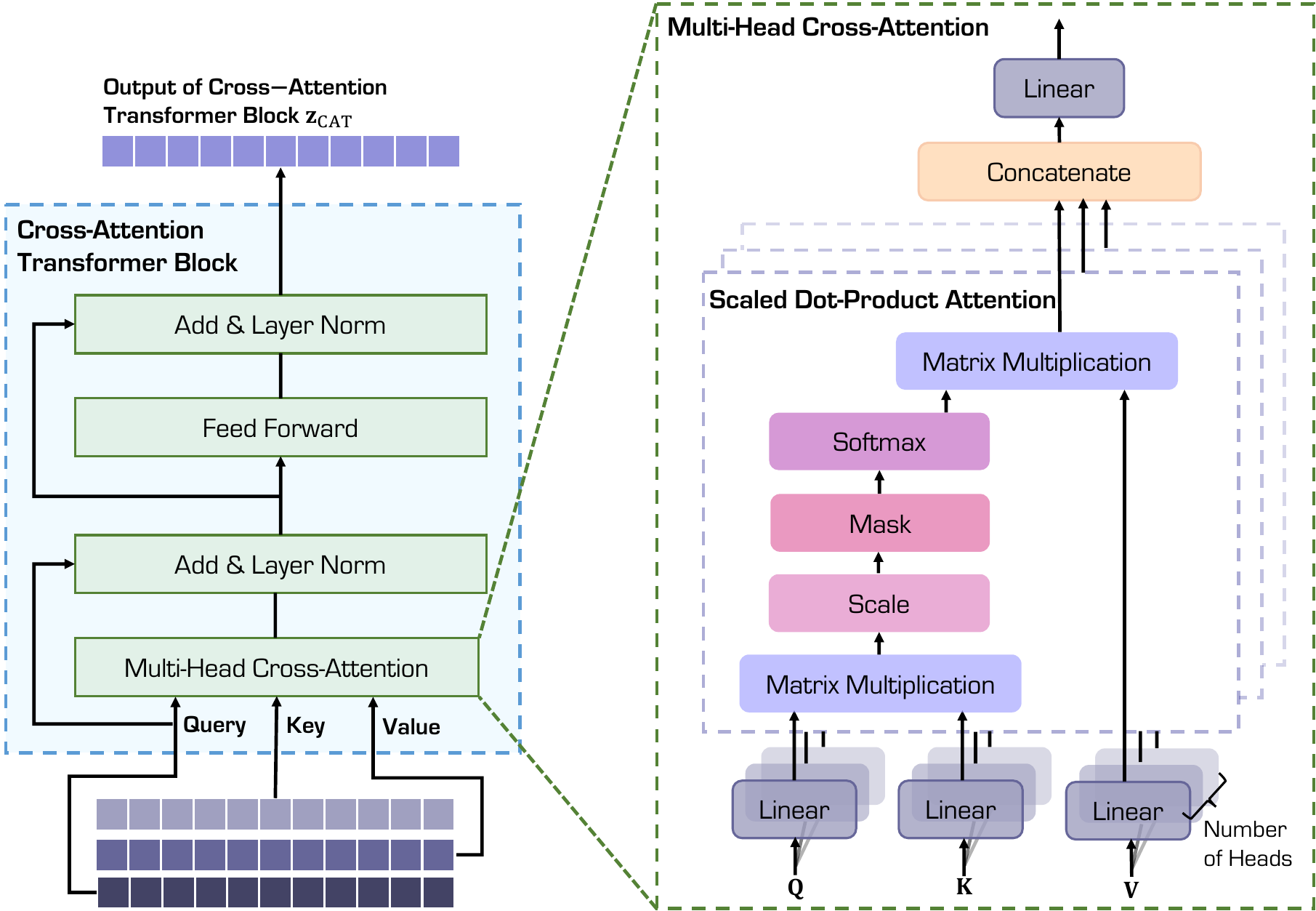}
\caption{Architecture of cross-attention transformer block for modeling car-following dependencies.}
\label{fig:yourlabel}
\end{figure}

\subsection{Denoising Network}

Fig. 5 shows the structure of the denoising network, which is responsible for reversing the diffusion process and reconstructing the trajectory. The network takes the noised future trajectory $\mathbf{x}_{{\rm fol}, K}^{\rm fut} \in \mathbb{R}^{T_{\text{fut}} \times D}$ and the interaction-enhanced output \(\mathbf{z}_{\text{CAT}}\) as inputs. During the downsampling stage, \(\mathbf{z}_{\text{CAT}}\) is concatenated with the feature maps after the first convolutional block to introduce vehicle interaction conditions. The network applies transposed convolutional layers during upsampling to progressively refine the trajectory. The final output is the predicted noise \(\hat{\epsilon}_{\theta} \in \mathbb{R}^{T_{\text{fut}} \times D}\), which is iteratively removed to recover the denoised trajectory.

\begin{figure}[htbp]
\centering
\includegraphics[width=.5\textwidth]{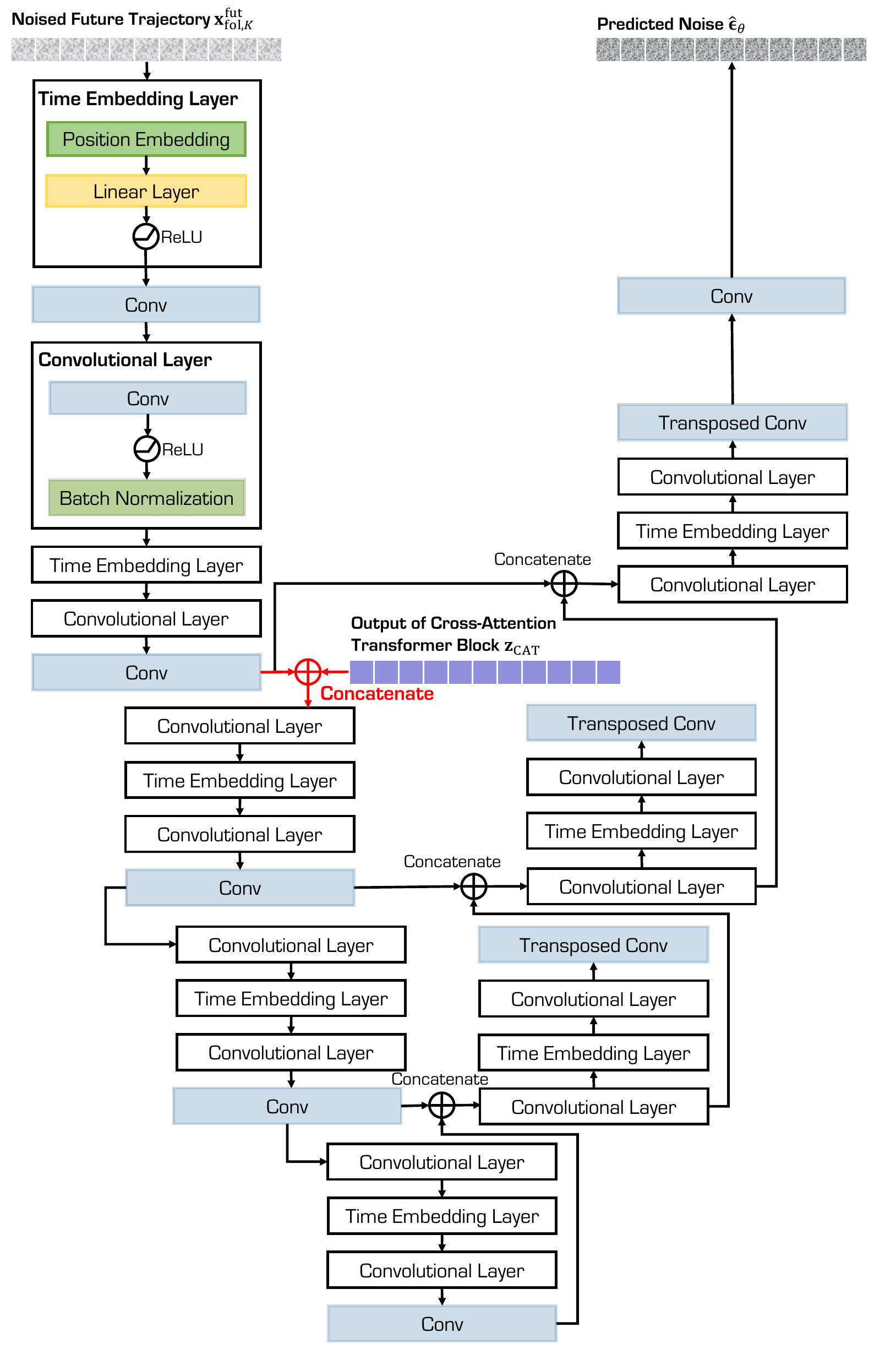}
\caption{Architecture of car-following condition incorporated U-Net \cite{ronneberger2015u} denoising network.}
\label{fig:yourlabel}
\end{figure}

\section{Implementation Details}

In addition to the implementation details provided in the main text, we present a more specific breakdown of the hyperparameters and configurations used in FollowGen, as listed in Table 4.

\begin{table}[ht]
  \centering
  \caption{Hyperparameters of FollowGen.}
  \begin{tabular}{>{\centering}m{5cm} c}
    \toprule[1.0pt]
    Parameter & Value \\
    \midrule
    Hidden Size of GRU  & 50 \\
    No. of GRU Layers   & 2 \\
    Embedding Size of Cross-Attention Transformer & 50 \\
    No. of Attention Heads & 5 \\
    Feed Forward Size & 100 \\
    No. of Down Sampling Channels in U-Net & (8, 16, 32, 64, 128) \\
    No. of Up Sampling Channels in U-Net & (128, 64, 32, 16, 8) \\
    $\beta_0$ & 0.0001 \\
    $\beta_K$ & 0.02 \\
    \bottomrule[1.0pt]
  \end{tabular}
\end{table}

\section{Additional Results}

While Fig. 3 in the main text visualizes the performance trends of the proposed FollowGen model under different diffusion steps $K$ using different metrics, Table 5 provides a more comprehensive and direct reference by reporting the quantitative results for all evaluation metrics across different scenarios and diffusion steps.

\begin{table*}
\centering
\caption{Performance comparison using different diffusion steps $K$.}
\resizebox{\textwidth}{!}{%
\begin{tabular}{c|c|cccc|cccc|cccc}
\hline
\multirow{3}{*}{\begin{tabular}[c]{@{}c@{}}Scenario\\ \\ \end{tabular}} & \multirow{3}{*}{\begin{tabular}[c]{@{}c@{}}Diffusion Steps $K$ \\ \\ \end{tabular}} & \multicolumn{4}{c|}{$T = 3s$} & \multicolumn{4}{c|}{$T = 3s$} & \multicolumn{4}{c}{$T = 3s$} \\ \cline{3-14} 
 &  & \textbf{RMSE $\downarrow$} & \textbf{ADE $\downarrow$} & \textbf{FDE $\downarrow$} & \textbf{MR $\downarrow$} & \textbf{RMSE $\downarrow$} & \textbf{ADE $\downarrow$} & \textbf{FDE $\downarrow$} & \textbf{MR $\downarrow$} & \textbf{RMSE $\downarrow$} & \textbf{ADE $\downarrow$} & \textbf{FDE $\downarrow$} & \textbf{MR $\downarrow$} \\ \hline
\multirow{4}{*}{H-H} & 50 & 16.1380 & 7.2097 & 13.4412 & 0.5182 & 21.2325 & 9.3392 & 17.5293 & 0.5643 & 26.0242 & 11.3947 & 21.3588 & 0.6497 \\ 
 & 100 & 8.0603 & 2.8001 & 4.0687 & \textbf{0.3464} & 9.9654 & 3.2994 & 5.2600 & 0.4107 & 12.1115 & 3.8830 & 7.4771 & 0.5560 \\ 
 & 500 & 3.8317 & 2.4699 & 2.7729 & 0.4721 & 4.1386 & 2.6424 & 3.3374 & 0.5240 & 4.6181 & 2.8892 & 5.3329 & 0.6073  \\ 
 & \textbf{200} & \textbf{2.8001} & \textbf{1.6162} & \textbf{2.1796} & 0.3820 & \textbf{3.3270} & \textbf{1.7993} & \textbf{2.4480} & \textbf{0.3930} & \textbf{3.8935} & \textbf{1.9853} & \textbf{3.3454} & \textbf{0.4935}  \\ \hline\hline
\multirow{4}{*}{A-H} & 50 & 14.7470 & 7.7627 & 12.8668 & 0.7938 & 16.9540 & 9.2661 & 15.4121 & 0.8391 & 23.5384 & 11.8312 & 74.3740 & 0.9531 \\ 
 & 100 & 10.0466 & 5.3686 & 7.2776 & 0.7438 & 10.4828 & 5.9673 & 8.4417 & 0.7750 & 12.4483 & 6.8167 & 20.1228 & 0.8359 \\ 
 & 500 & 2.9944 & 2.1644 & 2.1695 & 0.4359 & 2.9964 & 2.1902 & 2.3481 & 0.4766 & 3.3457 & 2.3913 & 5.8125 & 0.7203   \\ 
 & \textbf{200} & \textbf{2.0033} & \textbf{1.3058} & \textbf{1.3243} & \textbf{0.1891} & \textbf{2.1220} & \textbf{1.3585} & \textbf{1.5601} & \textbf{0.2547} & \textbf{2.4108} & \textbf{1.5058} & \textbf{3.3469} & \textbf{0.5750}  \\ \hline\hline
\multirow{4}{*}{H-A} & 50 & 10.5539 & 4.4520 & 8.8226 & 0.6582 & 14.1458 & 5.9700 & 11.9162 & 0.7069 & 18.9326 & 7.8506 & 25.5758 & 0.7153 \\  
 & 100 & 4.6471 & 1.9843 & 2.8179 & 0.3925 & 5.9825 & 2.6379 & 3.1814 & \textbf{0.3563} & 8.6185 & 2.6379 & 11.5274 & 0.6307 \\ 
 & 500 & 2.1538 & 1.5362 & 1.7423 & 0.3575 & 2.2948 & 1.6493 & 1.9785 & 0.3937 & 3.4587 & 1.8602 & 5.9887 & 0.6749 \\ 
 & \textbf{200} & \textbf{1.9550} & \textbf{1.3218} & \textbf{1.7257} & \textbf{0.3289} & \textbf{2.1989} & \textbf{1.4516} & \textbf{1.9596} & 0.3758 & \textbf{2.4810} & \textbf{1.5970} & \textbf{2.5775} & \textbf{0.4863} \\ \hline
\end{tabular}%
}
\end{table*}

From the table, we observe that the choice of diffusion steps $K$ significantly affects performance across all metrics and scenarios. Specifically, for $K = 200$, the model achieves a balance between accuracy and computational efficiency, as evidenced by the consistently lower values in RMSE, ADE, FDE, and MR across all scenarios. Increasing $K$ to 500 slightly improves RMSE and ADE in a few cases but at the cost of diminishing returns in FDE and MR. Conversely, a smaller $K$ leads to degraded performance across all metrics. These results highlight the importance of tuning $K$ to optimize FollowGen’s performance. As indicated in the main text, $K = 200$ offers an ideal tradeoff, which is consistent with its superior performance across the table.

\section{Visualization}

\begin{figure*}[htbp]
\centering
\includegraphics[width=\textwidth]{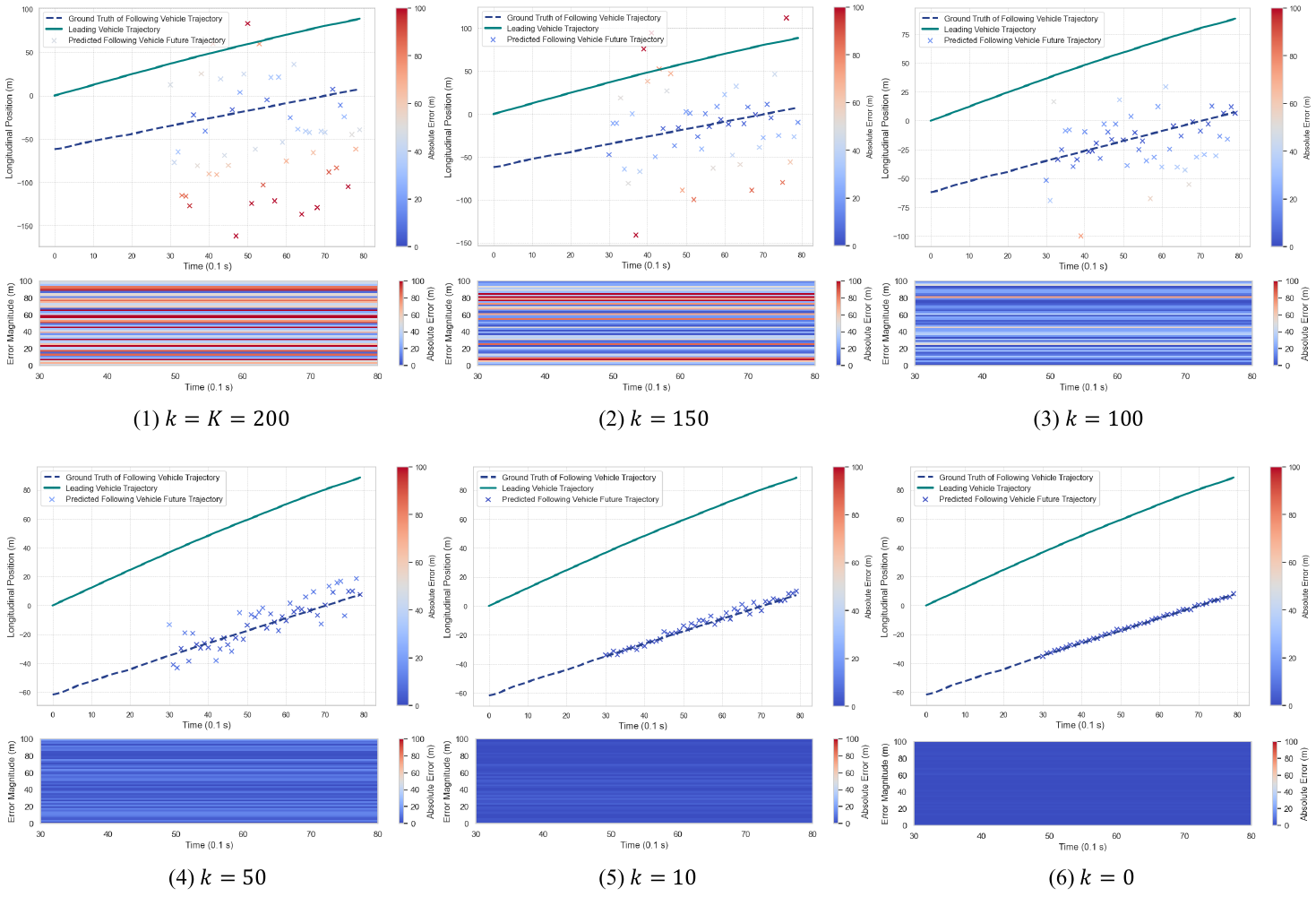}
\caption{Visualization of the sampling process in FollowGen's inference stage. The top row shows the predicted longitudinal positions of the following vehicle's trajectory (blue crosses), the ground truth trajectory (dashed line), and the leading vehicle's trajectory (solid line) across reverse time steps \(k\). Absolute errors are color-coded to reflect how close the recovered trajectory is to the ground truth. The bottom row visualizes the temporal evolution of error magnitude, with the color bar beneath each plot directly representing the degree of chaos at each step. The figure illustrates how chaotic scaled noise transitions into accurate trajectory predictions as the denoising progresses.}
\label{fig:yourlabel}
\end{figure*}

In this section, we visualize the sampling process, starting from completely chaotic scaled noise and progressively recovering the trajectory prediction. Fig. 6 illustrates an example of this process during the inference stage of FollowGen. The top row shows the evolution of the predicted longitudinal positions of the following vehicle's trajectory along the reverse time steps \(k\), alongside the ground truth trajectory, as well as the leading vehicle's trajectory. Absolute errors at each prediction step are marked using color coding, where the intensity of the color represents the magnitude of the error. This visualization demonstrates how the predicted trajectory transitions from the initial noisy state to align with the ground truth as noise is iteratively removed.

The bottom row visualizes the temporal progression of error metrics during the reverse time steps \(k\). The color bar beneath each plot directly illustrates the degree of chaos at each time step, reflecting the model's refinement of the trajectory prediction over time. As the denoising process progresses, the color intensity diminishes, indicating a reduction in error and a more coherent trajectory. These visualizations emphasize the effectiveness of FollowGen's condition-guided denoising procedure, which showcases its ability to handle chaotic initial noise and converge to precise trajectory predictions in car-following scenarios.

\end{document}